\documentclass[letterpaper, 10 pt, conference]{ieeeconf}  

\usepackage{times}
\usepackage{epsfig}
\usepackage{graphicx}
\usepackage{amsmath}
\usepackage{amssymb}
\usepackage{fontawesome}

\usepackage{tikz}
\usepackage{comment}
\usepackage[utf8]{inputenc} 
\usepackage[T1]{fontenc}    
\usepackage{url}            
\usepackage{amsfonts}       
\usepackage{nicefrac}       
\usepackage{microtype}      
\usepackage[bb=boondox]{mathalfa}
\usepackage{amsmath}
\usepackage{multirow}
\usepackage{algorithmic}
\usepackage{algorithm}
 \usepackage{wrapfig}
\usepackage{rotating}
\usepackage{hyperref}

\usepackage[utf8]{inputenc} 
\usepackage[T1]{fontenc}    
\usepackage{url}            
\usepackage{booktabs}       
\usepackage{amsfonts}       
\usepackage{nicefrac}       
\usepackage{microtype}      
\usepackage{marvosym}

\usepackage{float}
\usepackage{amssymb}
\usepackage{caption}
\usepackage[capitalize]{cleveref}

\usepackage{subcaption}
\crefname{section}{Sec.}{Secs.}
\Crefname{section}{Section}{Sections}
\Crefname{table}{Table}{Tables}
\crefname{table}{Tab.}{Tabs.}

\usepackage{xurl}

\IEEEoverridecommandlockouts                              

\title{\LARGE \bf 
JuggleRL: Mastering Ball Juggling with a Quadrotor via \\Deep Reinforcement Learning
}

\author{Shilong Ji$^{1}$, 
Yinuo Chen$^{1}$,
Chuqi Wang$^{1}$,
Jiayu Chen$^{1}$,
Ruize Zhang$^{1}$,
Feng Gao$^{1}$,
Wenhao Tang$^{1}$,\\
Shu'ang Yu$^{1}$,
Sirui Xiang$^{1}$,
Xinlei Chen$^{1}$,
Chao Yu$^{1}$\textsuperscript{\Letter},
Yu Wang$^{1}$\textsuperscript{\Letter}
\thanks{{\Letter} Corresponding Authors. \tt{\{yuchao,yu-wang\}@tsinghua.edu.cn}}
\thanks{$^{1}$ Tsinghua University, Beijing, 100084, China.}%
\thanks{This research was supported by National Natural Science Foundation of China (No.62406159, 62325405), Postdoctoral Fellowship Program of CPSF under Grant Number (GZC20240830, 2024M761676), China Postdoctoral Science Special Foundation 2024T170496.}
}

\begin{document}

\maketitle
\thispagestyle{empty}
\pagestyle{empty}


\begin{abstract}
Aerial robots interacting with objects must perform precise, contact-rich maneuvers under uncertainty. In this paper, we study the problem of aerial ball juggling using a quadrotor equipped with a racket, a task that demands accurate timing, stable control, and continuous adaptation. We propose JuggleRL, the first reinforcement learning-based system for aerial juggling. It learns closed-loop policies in large-scale simulation using systematic calibration of quadrotor and ball dynamics to reduce the sim-to-real gap. The training incorporates reward shaping to encourage racket-centered hits and sustained juggling, as well as domain randomization over ball position and coefficient of restitution to enhance robustness and transferability. The learned policy outputs mid-level commands executed by a low-level controller and is deployed zero-shot on real hardware, where an enhanced perception module with a lightweight communication protocol reduces delays in high-frequency state estimation and ensures real-time control. Experiments show that JuggleRL achieves an average of $311$ hits over $10$ consecutive trials in the real world, with a maximum of $462$ hits observed, far exceeding a model-based baseline that reaches at most $14$ hits with an average of $3.1$. Moreover, the policy generalizes to unseen conditions, successfully juggling a lighter $5$ g ball with an average of $145.9$ hits. This work demonstrates that reinforcement learning can empower aerial robots with robust and stable control in dynamic interaction tasks.
\end{abstract}

\section{Introduction}
\label{sec:intro}

Dynamic interaction with the environment remains one of the central challenges in robotics. Tasks such as pushing, striking, or juggling are inherently reactive, contact-rich, and must be performed under tight temporal constraints, often in the presence of model uncertainty, sensing noise, and environmental variability. Aerial robots face these challenges even more acutely, as they must not only stabilize themselves in flight but also coordinate precise interactions with fast-moving objects. In this paper, we study the problem of aerial ball juggling using a quadrotor equipped with a racket. Despite its simplicity, this task demands accurate timing, robust contact handling, and rapid adaptation, making it a representative benchmark for real-time interaction under uncertainty.

Traditional control approaches~\cite{eth, eth1, sjtu} address this problem by predicting ball trajectories, calculating contact points, and planning quadrotor motions to satisfy these constraints. However, such methods are brittle in practice, due to several fundamental limitations:
\begin{itemize}
    \item \textbf{Nonlinear ball dynamics}: The ball’s motion model is highly nonlinear and lacks a closed-form solution. Existing methods rely on numerical integration, which inevitably trades off between accuracy and speed. Such computational limitations introduce timing errors that become especially critical under low-bounce conditions, often driving the system into physically infeasible states.
    \item \textbf{Error accumulation}: Trajectory planning must be performed in advance based on predicted contact points. Because subsequent execution follows an open-loop strategy without incorporating new observations, any prediction error cannot be corrected and instead accumulates over time, ultimately leading to a severe degradation in task success rates.
    \item \textbf{Hard trajectory constraints}: Trajectory optimization requires predefined terminal states and hitting times. These hard constraints shrink the feasible solution space and often lead to infeasibility under real-world uncertainties.
\end{itemize}

Reinforcement learning (RL), in contrast, offers a compelling alternative paradigm. Rather than explicitly computing optimal trajectories or solving inverse dynamics during execution, RL acquires sensorimotor policies through interaction, thereby enabling closed-loop feedback, online error correction, and adaptive responses to novel disturbances.

In this paper, we present \textbf{JuggleRL}, the first reinforcement learning–based system for aerial ball juggling. The framework is developed in three stages. First, we perform system identification (SysID) to calibrate the dynamics of both the quadrotor and the ball, narrowing the gap between the physical platform and its digital counterpart. Second, we construct a large-scale GPU-parallel training environment based on Isaac Sim~\cite{gym,orbit,isaacsim}, supporting contact-rich interactions. Within this environment, policies are trained using Proximal Policy Optimization (PPO)~\cite{ppo}. The training incorporates reward shaping to encourage racket-centered impacts and sustained juggling, as well as domain randomization over ball position, coefficient of restitution, and other uncertain parameters to improve robustness and sim-to-real transfer. The policy outputs mid-level collective thrust and body rates (CTBR) commands, which are executed by a low-level PID controller on the quadrotor. Finally, the trained policy is deployed zero-shot on real hardware, where an enhanced perception module, incorporating a lightweight communication protocol, reduces latency in high-frequency state estimation, ensuring real-time control.

Extensive experiments validate the effectiveness of JuggleRL. In simulation, we compared JuggleRL with a traditional model-based predictive planner under varying ball drop heights. JuggleRL consistently maintained juggling performance close to the maximum episode length across all heights, whereas the baseline failed whenever the drop height fell below $4$m. In real-world experiments, our policy achieved up to $462$ consecutive hits, with an average of $311$ hits over $10$ trials, while the baseline reached at most $14$ hits and averaged only $3.1$ hits. Moreover, the learned policy generalizes to previously unseen ball weights, successfully juggling a lighter $5$g ball with an average of $145.9$ hits over $10$ consecutive trials, demonstrating robust performance and adaptability to novel physical conditions.

Our contributions can be summarized as follows:
\begin{itemize}
	\item We present the first model-free RL-based aerial ball-juggling system, JuggleRL, with zero-shot sim-to-real deployment.
    \item JuggleRL integrates a series of techniques, including system identification, soft-constraint reward shaping, domain randomization, and a lightweight communication protocol in the perception module to ensure robust real-time control on real hardware.
	\item JuggleRL achieves up to $462$ hits with an average of $311$ in real experiments under varying ball drop heights, far exceeding the baseline, which reaches at most $14$ hits with an average of $3.1$.
	\item JuggleRL could generalize to unseen ball weights, successfully juggling a lighter $5$g ball with an average of $145.9$ hits over $10$ consecutive trials.
\end{itemize}
\section{Related Work}
\subsection{Motion Planning}
Motion planning and control for autonomous quadrotors have traditionally followed two major paradigms: optimization-based methods and reinforcement learning (RL).
Optimization-based methods decompose the problem into path planning and trajectory optimization, producing smooth and executable trajectories. This line of research has evolved from the seminal Minimum Snap trajectory generation~\cite{minsnap} to advanced formulations such as Minimum Jerk planning~\cite{minjerk}, polynomial-based navigation in dense environments~\cite{poly}, and gradient-based approaches that avoid explicit signed-distance fields~\cite{egoplanner}, with further extensions to decentralized swarm flight~\cite{egoswarm}. While these methods offer strong guarantees on trajectory feasibility, they often involve heavy online computation~\cite{flightbench}.

In contrast, RL provides an end-to-end alternative, directly mapping sensory inputs to control commands. Policies are typically trained in high-fidelity simulators~\cite{goggles,omnidrones} and executed with high inference efficiency in the real world. RL has enabled quadrotors to achieve impressive capabilities, including stable hovering under disturbances~\cite{hwangbo2017control}, minimum-time navigation in cluttered environments~\cite{penicka2022learning}, superhuman-level drone racing~\cite{kaufmann2023champion}, and aggressive flight maneuvers~\cite{sun2022aggressive}.

Despite their differences, both paradigms have largely concentrated on free-flight control, where the quadrotor operates without physical interaction with external objects.
In this paper, we investigate a fundamentally different problem: aerial ball juggling. Unlike free-flight tasks, juggling requires sustained dynamic interaction between the quadrotor and a ball in motion. The vehicle must execute precise spatio-temporal strikes—arriving at specific positions with appropriate velocities and orientations—while continuously adapting to the ball’s real-time dynamics. These requirements—precise timing, rapid response, and robustness to sensing and communication delays—make aerial ball juggling far more demanding than conventional flight control, where traditional model-based methods achieve at most $14$ consecutive hits~\cite{eth,sjtu}. By contrast, our RL-based system succeeds in achieving up to $462$ consecutive hits, demonstrating that reinforcement learning can enable robust, adaptive, and interactive quadrotor control.

\subsection{Aerial Ball-juggling}
Previous work has explored quadrotor-based ball juggling primarily through prediction–planning frameworks.
Müller et al. \cite{eth} proposed a seminal system that predicts ball states with a Kalman filter, models impact dynamics with a restitution-based rigid body model, and solves an open-loop interception trajectory for the quadrotor. This paradigm has since become the foundation for subsequent research.
Ritz et al. \cite{eth1} improved the planning module by optimizing trajectories with a minimum-acceleration criterion, enabling smoother and more flyable paths for cooperative juggling with two quadrotors. Dong et al. \cite{sjtu} extended the framework by introducing minimum-jerk reference trajectories \cite{mueller2013computationally} and robust feedback control, addressing under-actuation and enhancing efficiency. Yu et al.~\cite{batplanner} further refined the pipeline by employing an extended Kalman filter (EKF), cubic spline interpolation for trajectory fitting, and a two-stage joint optimization method for contact-aware planning.

Despite these advances, prediction–planning approaches face fundamental limitations.
The ball’s motion exhibits strong nonlinearities (e.g., aerodynamic drag) that are difficult to model accurately. Reliable optimization at contact requires manually defined heuristics due to incomplete or uncertain state information. Moreover, online re-planning often introduces abrupt trajectory shifts, while pre-computed open-loop trajectories inevitably accumulate prediction errors. Collectively, these challenges constrain the robustness and success rate of existing juggling systems.

In contrast, our proposed RL–based system directly optimizes the final objective—the number of successful hits—without relying on predefined intermediate states.
Rather than predicting long-term ball trajectories, our method only filters the instantaneous ball state as input. The learned policy updates quadrotor actions in real time, implicitly achieving adaptive motion planning and reducing reliance on handcrafted models. This end-to-end approach enables stable, robust juggling under diverse and uncertain conditions.

\section{Preliminary}
\subsection{Problem Formulation}

We model the aerial ball-juggling task as a Markov Decision Process, characterized by $M = \langle\mathcal{S}, \mathcal{A}, \mathcal{O}, \mathcal{P}, \mathcal{R}, \gamma\rangle$, which defines the state space $\mathcal{S}$, action space $\mathcal{A}$, and observation space $\mathcal{O}$, the transition probability $\mathcal{P}$, reward function $\mathcal{R}$, and discount factor $\gamma$. The objective is to learn a stochastic policy $\pi_\theta(\mathbf{a}_t|\mathbf{o}_t)$, parameterized by $\theta$, that maps the current observation $\mathbf{o}_t $ to a distribution over actions $\mathbf{a}_t$, with the aim of achieving \textit{sustained ball juggling}. The parameters $\theta$ are optimized to maximize the accumulative reward $J(\theta)= \mathbb{E}_{\pi_\theta} \left[\sum_t \gamma^t R(\mathbf{s}_t, \mathbf{a}_t)\right]$.

\begin{figure*}[t]
\centering
\includegraphics[width=1.0\textwidth]{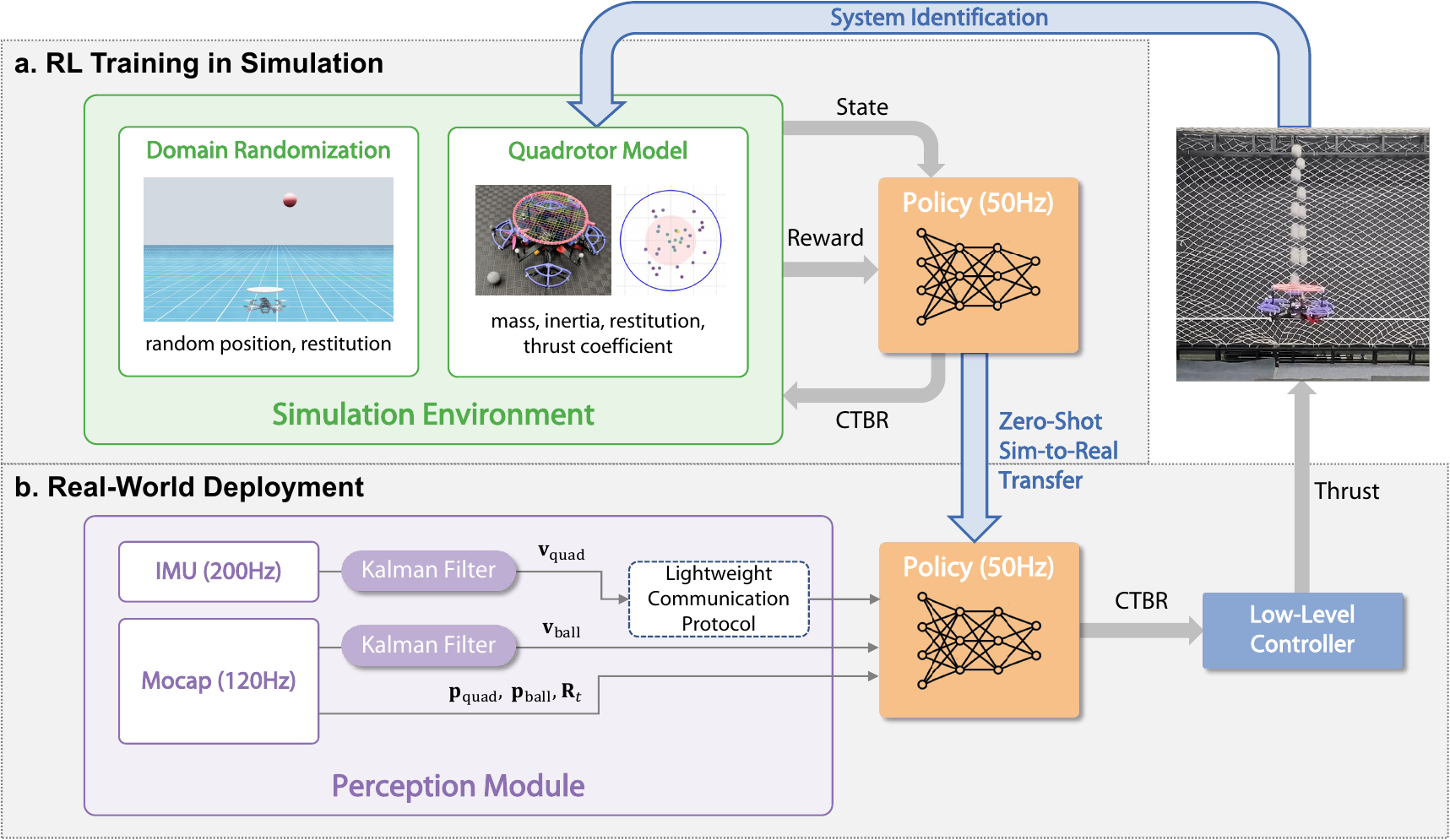}
\caption{Overview of the JuggleRL system. The system combines SysID-based dynamics calibration, large-scale GPU-parallel training in Isaac Sim with domain randomization to learn robust juggling policies. The learned policy outputs mid-level CTBR commands executed by a low-level PID controller, and is deployed zero-shot on real hardware with an enhanced perception module using a lightweight communication protocol to minimize latency in high-frequency state estimation.}
\label{fig:overview}
\end{figure*}

\subsection{Quadrotor Dynamics}
A quadrotor is modelled as a $6$ degree-of-freedom rigid body of mass $m$ and inertia matrix $\mathbf{I} = [\mathbf{I}_x, \mathbf{I}_y, \mathbf{I}_z]$. Its dynamics are modelled by: 
\begin{equation}
    \mathbf{\dot{x}} = 
    \begin{bmatrix}
    \mathbf{\dot{p}}_{\mathcal{W}} \\
    \mathbf{\dot{q}} \\
    \mathbf{\dot{v}}_{\mathcal{W}} \\
    \mathbf{\dot{\omega}}_{\mathcal{B}}\\
    
    \mathbf{\dot{\Omega}}
    \end{bmatrix}
    = 
    \begin{bmatrix}
    \mathbf{v}_{\mathcal{W}} \\
    \mathbf{q}\textcircled{\boldmath$\times$}[0, \mathbf{\omega}_{\mathcal{B}}/2]^T\\
    \frac{1}{m}\mathbf{q}\cdot\mathbf{f}_{\text{prop}}\cdot\overline{\mathbf{q}} + \mathbf{g}_{\mathcal{W}} \\
    \mathbf{I}^{-1}(\mathbf{\tau}_{\text{prop}} - \mathbf{\omega}_{\mathcal{B}}\times(\mathbf{I}\mathbf{\omega}_{\mathcal{B}})) \\
    T_m(\mathbf{\Omega}_{\text{cmd}} - \mathbf{\Omega})
    \end{bmatrix},
\end{equation}
where state $\mathbf{x}$ consists of position $\mathbf{p}$, quaternion $\mathbf{q}$, linear velocity $\mathbf{v}$, angular velocity $\mathbf{\omega}$ and rotation speed of rotors $\mathbf{\Omega}$. Subscripts $\mathcal{W}$ and $\mathcal{B}$ represent the world and body frame. Frame $\mathcal{B}$ is located at the center of gravity. Notation $\textcircled{\boldmath$\times$} $ indicates the multiplication of two quaternions. $\overline{\mathbf q}$ denotes the quaternion’s conjugate. $\mathbf{g}_{\mathcal{W}} = [0, 0, -9.81m/s^2]^T$ denotes gravity. $\mathbf{f}_{\text{prop}}$ and $\mathbf{\tau}_{\text{prop}}$ are the collective force and the torque produced by the propellers, defined as:
\begin{equation}
    \mathbf{f}_{\text{prop}} = \Sigma_j\mathbf{f}_j, \mathbf{\tau}_{\text{prop}} = \Sigma_j\mathbf{\tau}_j + \mathbf{r}_{p, j}\times\mathbf{f}_j,
\end{equation}
where $\mathbf{r}_{p, j}$ is the location of propeller $j$ in the body frame, $\mathbf{f}_j, \mathbf{\tau}_j$ the force and torque of propeller $j$. The dynamic of $\mathbf{\Omega}$ is modelled as a first-order system with a time constant $T_m$. 
We adopt a widely used model from prior work for the calculation of force and torque. 
Each motor's thrust and torque are modelled as proportional to the square of its rotation speed~\cite{furrer2016rotors}, determined by thrust coefficient $k_f$ and drag coefficient $k_m$:
\begin{equation}
    \mathbf{f}_j = [0, 0, k_f\Omega_j^2]^T, \mathbf{\tau}_j = [0, 0, k_m\Omega_j^2]^T.
\end{equation}

\subsection{Task Setup}

We study the problem of \textit{aerial ball juggling}, where a quadrotor equipped with a badminton racket must repeatedly strike a ball to keep it aloft. The task objective is to maximize the number of consecutive successful hits before failure.  

At each time step, the agent observes both the quadrotor state and the ball state, obtained from simulation or onboard estimation. A hit is considered successful only if two spatial constraints are satisfied: the contact height lies within a prescribed interval, and the apex of the subsequent trajectory also falls within a valid range, namely
\begin{subequations}\label{eq:task-constraints}
\begin{align}
z_{\min}^{\text{hit}} \le z_{\text{hit}} \le z_{\max}^{\text{hit}}, \label{eq:hit-window}\\
z_{\min}^{\text{apex}} \le z_{\text{apex}} \le z_{\max}^{\text{apex}}. \label{eq:apex-window}
\end{align}
\end{subequations}
These constraints ensure that each strike produces a controllable rebound and that the ball remains within the workspace. The episode terminates once the ball touches the ground or leaves the workspace.  

This setup imposes tight spatio--temporal requirements: the quadrotor must position itself precisely to satisfy the contact window (see Equ.~\eqref{eq:hit-window}) while ensuring the next apex remains feasible
(see Equ.~\eqref{eq:apex-window}). Even small delays or misalignments quickly lead to failure, reflecting the difficulty of executing accurate strikes with an underactuated platform subject to strict timing constraints.

\section{JuggleRL System}


\subsection{Overview}
The overview of the JuggleRL system is shown in Fig.~\ref{fig:overview}. The framework begins with system identification (SysID) to calibrate the dynamics of both the quadrotor and the ball, ensuring consistency between the physical platform and its digital counterpart. We then construct a large-scale training environment, where policies are trained using PPO~\cite{ppo}. To improve robustness and facilitate sim-to-real transfer, reward shaping and domain randomization are incorporated into training. Finally, for real-world deployment, we design an enhanced perception module equipped with a lightweight communication protocol, which reduces latency in high-frequency state estimation and enables robust real-time control on physical quadrotors. 

\subsection{System Identification}
\begin{figure}[t]
\centering
\includegraphics[width=0.8\columnwidth]{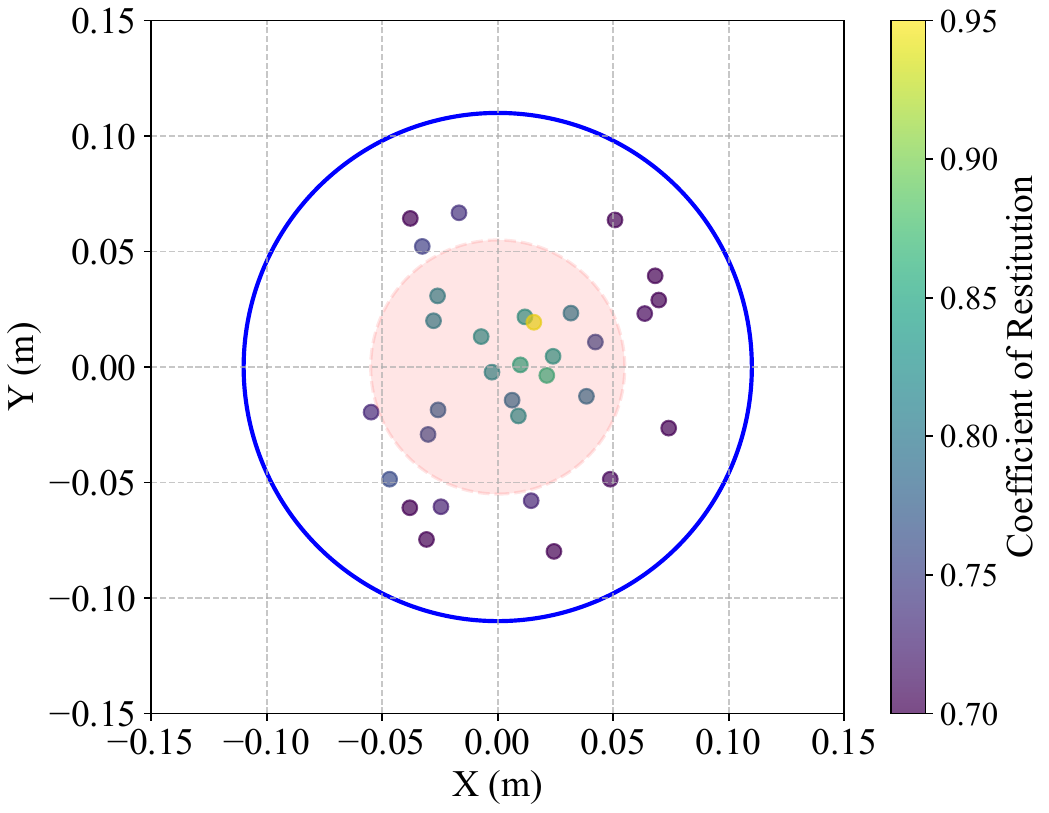}
\caption{Visualization of the restitution coefficient between the racket and the ball. The central "sweet spot" exhibits a higher restitution coefficient (average $0.82$), while the periphery is less elastic (average $0.64$), motivating our domain randomization strategy.}
\label{fig:paddle_elasticity} 
\end{figure}

We precisely measure the key dynamic parameters of our quadrotor through SysID to construct a high-fidelity simulation model. As summarized in Tab.~\ref{tab:sysid_params}, these parameters include the quadrotor’s mass $m$, the inertia $\mathbf{I} = [\mathbf{I}_x, \mathbf{I}_y, \mathbf{I}_z]$, the motor thrust coefficient $k_f$, as well as the mass of the ball $m_{\text{ball}}$ and the coefficient of restitution between the racket surface and the ball $R_s$. The mass $m$ and inertia $\mathbf{I}$ of the quadrotor, along with the mass of the ball $m_{\text{ball}}$, are measured directly using physical methods. The motor thrust coefficient $k_f$ is estimated by first calculating the maximum thrust of a single motor from the hover throttle and the quadrotor's mass, and then normalizing the thrust by the square of the maximum motor angular velocity.
To determine the coefficient of restitution $R_s$ between the racket and the ball, we hover the quadrotor and racket stationary in the air and measure the rebound behavior of the ball dropped freely onto the racket at different contact points. 
A region is defined as the ``sweet spot'' if its coefficient of restitution remains above 0.75 throughout the area; otherwise, it is regarded as non-sweet.
As shown in Fig.~\ref{fig:paddle_elasticity}, the central circular region of the racket, with a radius of $5.5$ cm, can be considered the ``sweet spot,'' exhibiting an average coefficient of restitution of $0.82$, while the outer regions have a lower average of approximately $0.64$.
\begin{table}[t]
\caption{Parameters of the quadrotor and the ball}
\label{tab:sysid_params}
\centering
\begin{tabular}{l l}
\toprule
\textbf{Parameter} & \textbf{Value} \\
\midrule
$m$  ($kg$)  & $1.090$ \\
$\mathbf{I}_x$ ($kg \cdot m^2$) & $[5.29 \times 10^{-3}, 0, 0]$  \\
$\mathbf{I}_y$ ($kg \cdot m^2$) & $[0, 5.62 \times 10^{-3}, 0]$ \\
$\mathbf{I}_z$ ($kg \cdot m^2$) & $[0, 0, 7.80 \times 10^{-3}]$ \\
$k_f$ ($kg\cdot m$) & $5.39 \times 10^{-6}$ \\
$m_{\text{ball}}$  ($kg$)  & $0.0472$ \\
$R_s$ & $0.82$ (sweet), $0.64$ (outer) \\
\bottomrule
\end{tabular}
\end{table}

\subsection{Reinforcement Learning with Generalization Enhancement}
We use PPO~\cite{ppo} as the training algorithm and provide a detailed design of the observation space, action space, reward function, and domain randomization below.

\textbf{Observation Space.}
The quadrotor's input consists of two components: a state space observable by both the actor and critic networks, and privileged information accessible only to the critic.
The $24$-dimensional state space includes the quadrotor's absolute root state $\mathbf{o}_{\text{quad},t}=[\mathbf{p}_{\text{quad},t}, \mathbf{R}_t, \mathbf{v}_{\text{quad},t}] \in \mathbb{R}^{15}$, where $\mathbf{p}_{\text{quad},t}\in \mathbb{R}^{3}$, $\mathbf{v}_{\text{quad},t}\in \mathbb{R}^{3}$ and $\mathbf{R}_t\in \mathbb{R}^{9}$ denote the position, linear velocity and flattened rotation matrix of the quadropter at time step $t$, respectively. In addition, the ball's observation vector is
$\mathbf{o}_{\text{ball},t} = [\mathbf{p}_{\text{ball},t}, \mathbf{v}_{\text{ball},t}, \mathbf{\Delta p}] \in \mathbb{R}^{9}$, where $\mathbf{\Delta p}\in \mathbb{R}^{3}$ denotes the relative position between the ball and the quadrotor. The privileged information for the critic is the current timestep $t$ which helps the critic network capture temporal dynamics for more accurate value estimation.

\textbf{Action Space.}
The action space is $4$-dimensional, representing the collective thrust and body rates (CTBR). The action at time step $t$ is defined as $\mathbf{a}_t = [c, \omega_x, \omega_y, \omega_z] \in \mathbb{R}^4$, where $c$ denotes the collective thrust controlling the quadrotor's vertical acceleration, and $(\omega_x, \omega_y, \omega_z)$ are the body rates around the roll, pitch, and yaw axes, respectively. 



\textbf{Reward Function.}
Our reward function combines sparse task rewards with several shaping terms. The core rewards are a hitting reward $r^{\text{hit}}$ for successful strikes within a specified altitude window (Equ.~\eqref{eq:hit-window}) and an apex reward $r^{\text{apex}}$ for reaching the desired apex height (Equ.~\eqref{eq:apex-window}). Shaping rewards include a dense positional term $r^{\text{rpos}}$, a center-contact reward $r^{\text{contact}}$, a horizontal position penalty $r^{\text{xy}}$, a smoothness reward $r^{\text{smooth}}$, and a spin reward $r^{\text{spin}}$.  

The reward at time $t$ is defined as:
\begin{subequations}\label{eq:reward}
\begin{align}
r^{\text{hit}}_t &= 50 \,\mathbb{1}_{\text{hit}}, \label{eq:reward-hit}\\
r^{\text{apex}}_t &= 50 \,\mathbb{1}_{\text{apex}}, \label{eq:reward-apex}\\
r^{\text{rpos}}_t &= \frac{1}{1 + \max(d_{xy},\,0.2)}, \label{eq:reward-rpos}\\
r^{\text{contact}}_t &= 10 \,\exp(-5\, d_{\text{axis}})\,\mathbb{1}_{\text{hit}}, \label{eq:reward-contact}\\
r^{\text{xy}}_t &= -1.0 \, d_{\text{quad}}, \label{eq:reward-xy}\\
r^{\text{smooth}}_t &= 2.0 \,\exp(-\varepsilon_t), \label{eq:reward-smooth}\\
r^{\text{spin}}_t &= -10 \, \lvert \psi_t \rvert, \label{eq:reward-spin}
\end{align}
\end{subequations}
where $d_{xy}$ is the horizontal distance between the quadrotor and the ball, $d_{\text{axis}}$ is the distance of the contact point from the racket centerline, $d_{\text{quad}}$ is the horizontal deviation of the quadrotor’s position from the origin, 
$\psi_t$ is the yaw angle of the quadrotor,
and $\varepsilon_t$ is the action error defined as:
\begin{equation}
\varepsilon_t = \| \mathbf{a}_{t} - \mathbf{a}_{t-1} \|_2,
\label{eq:action_smoothness}
\end{equation}

where $\mathbf{a}_{t}$ is the current action and $\mathbf{a}_{t-1}$ is the previous action. The indicator functions $\mathbb{1}_{\text{hit}}$ and $\mathbb{1}_{\text{apex}}$ are triggered by the conditions in Equ.~\eqref{eq:hit-window}--\eqref{eq:apex-window}.  

The overall reward is the sum of all components,
\begin{equation}
r_t = r^{\text{hit}}_t + r^{\text{apex}}_t + r^{\text{rpos}}_t + r^{\text{contact}}_t + r^{\text{xy}}_t + r^{\text{smooth}}_t + r^{\text{spin}}_t.
\end{equation}

\textbf{Domain Randomization.}
Tab.~\ref{tab:dr_params} summarizes our domain randomization settings including coefficient of restitution $R_s$, initial position of the quadrotor $X_{\text{quad0}}$, $Y_{\text{quad0}}$, $Z_{\text{quad0}}$, and initial position of the ball $X_{\text{ball0}}$, $Y_{\text{ball0}}$, $Z_{\text{ball0}}$. The height randomization aims to train a policy capable of stabilizing the ball from various initial heights, while the horizontal position randomization is employed to facilitate the policy in learning fast and precise actions to intercept balls with horizontal velocity.
Simulating the racket's central ``sweet pot'', we randomize the coefficient of restitution around the average value $0.82$ in the ``sweet pot'' and choose the range of $[0.75,0.90]$.

\begin{table}[t]
\caption{Range of domain randomization parameters for the quadrotor and ball.}
\label{tab:dr_params}
\centering
\begin{tabular}{l l} 
\toprule
\textbf{Parameters} & \textbf{Range} \\
\midrule
$R_s$ & $[0.75, 0.90]$ \\
$Z_{\text{ball0}}$ & $[1.5, 2.0]$ \\
$Y_{\text{quad0}}$ & $[0.9, 1.1]$ \\
$X_{\text{ball0}}$, $X_{\text{quad0}}$ & $[-0.07, 0.07]$ \\
$Y_{\text{ball0}}$, $Y_{\text{quad0}}$ & $[-0.07, 0.07]$ \\
\bottomrule
\end{tabular}
\end{table}

\subsection{Real-world Deployment}
For the real-world experiment, the reinforcement learning policy is zero-shot deployed onboard an NVIDIA Jetson Orin NX carried by the quadrotor platform. The policy takes both the quadrotor and ball states as input. The quadrotor poses are obtained from a motion capture system at $120$ Hz, while velocities are estimated on the onboard computer by fusing mocap and IMU measurements at $200$ Hz via a Kalman filter. The ball is tracked as a reflective marker in the Mocap system, and its velocity is recovered through Kalman filtering of position measurements.

Because these high-frequency state estimates, especially the $200$ Hz quadrotor velocities, incur heavy communication overhead and serialization delays, we design a lightweight communication protocol (LCP) that transmits only the essential motion states, i.e., the quadrotor’s pose and twist. This eliminates latency accumulation and ensures that the policy receives state updates in real time. Running at $50$ Hz, the policy then outputs CTBR, which are executed by the onboard low-level flight controller.

\begin{figure}[t]
\begin{minipage}{0.24\textwidth}
\centering
\subcaptionbox{without LCP.\label{fig:ab}}
{\includegraphics[width=\textwidth]{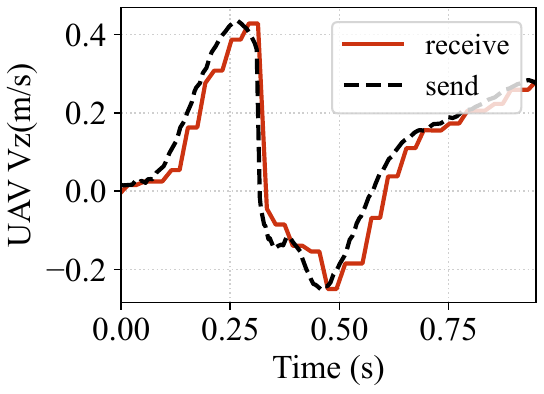}}
\end{minipage}
\begin{minipage}{0.24\textwidth}
\subcaptionbox{with LCP. \label{fig:speed}}{\includegraphics[width=\textwidth]{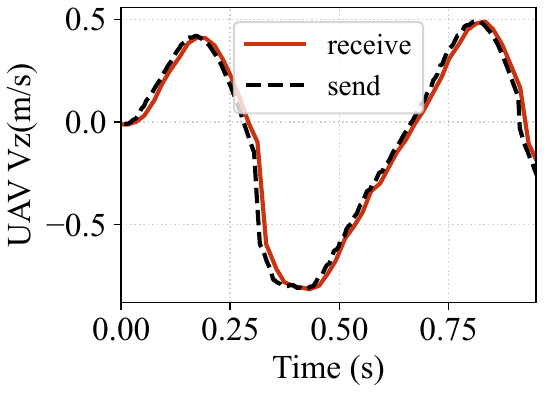}}
\end{minipage}
\caption{Latency comparison with the lightweight communication protocol (LCP). Using the original communication scheme, the velocity signal exhibits step-like latency even at a 200 Hz publish rate. In contrast, LCP enables smooth, real-time transmission of velocity.}
\label{fig:lcp}

\end{figure}

\section{Experiments}
\subsection{Experiment Setting}
We employ OmniDrones \cite{omnidrones}, a high-parallel UAV simulator, to construct the ball juggling task. The setup consists of a quadrotor (mass $1090$ g) juggling a ball (mass $47.2$ g, radius $2$ cm) falling from above, using a racket with a $5.5$ cm radius. Each episode is capped at $500$ steps. Training is conducted with $4096$ parallel environments running at $50$ Hz, producing $2$ billion simulation frames in under $3$ hours on a desktop workstation (Intel i7-13700K CPU, RTX $4090$ GPU).


\subsection{Evaluation Metric}
We adopt the \emph{number of consecutive hits} as the primary evaluation metric for juggling performance. In simulation, where conditions are idealized, the metric is defined as the maximum number of hits achieved within a fixed one-minute episode. Reported results are averaged over $100$ episodes for statistical reliability. In real-world experiments, the quadrotor continues juggling until failure, and we report the maximum number of hits over $10$ consecutive trials as an indicator of robustness.

\subsection{Baseline}
We adopt the open-source model-based juggling baseline from Müller et al.~\cite{eth}, referred to as the model-based predictive planner (MBPP). MBPP predicts the ball trajectory with a Kalman filter, generates an open-loop interception trajectory, and executes it through a trajectory tracking controller.

\subsection{Main Results}

\begin{figure}[htbp]
\centering
\includegraphics[width=0.9\columnwidth]{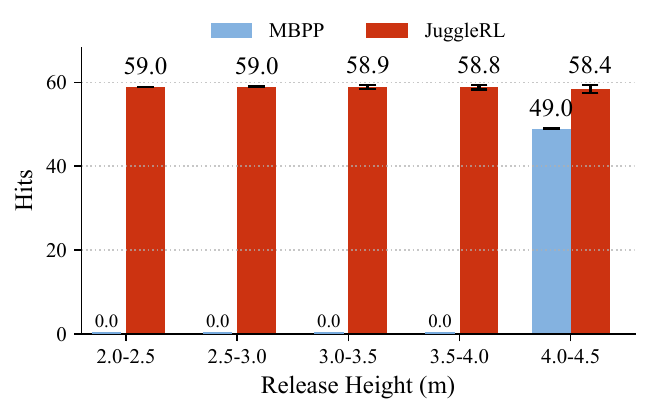}
\caption{Performance comparison between JuggleRL and MBPP in simulation. JuggleRL demonstrates robust performance across all tested ball release heights, while MBPP fails at lower release heights.}
\label{fig:sim_comparison}
\end{figure}

\textbf{Simulation Results.  } 
We first evaluate the performance of JuggleRL against MBPP in simulation. Fig.~\ref{fig:sim_comparison} compares different initial ball release heights. MBPP requires a fixed target apex height (set to $3$ m in our experiments) in order to compute a feasible trajectory, whereas JuggleRL operates within a target apex height range of $[2.85,3.2]$ m without such constraints. MBPP succeeds only when the ball is released between $4.0$ and $4.5$ m, and fails completely at lower heights, primarily because it relies on sufficient reaction time for the prediction–planning pipeline. In contrast, JuggleRL consistently maintains near-maximum episode performance (over $58$ hits) across all tested heights, demonstrating its reactive capability and robustness.

\textbf{Real-world Results. } We further evaluated JuggleRL in real-world experiments under two settings: fixed height, where the ball was released from $1.65$ m, and varying height, where the release height was randomized between $1.5$ m and $1.95$ m. The target apex height range is $[1.6,1.85]$ m.
Tab.~\ref{tab:real_world_main} summarizes the results over $10$ consecutive trials: under the fixed-height setting, JuggleRL achieved an average of $311$ consecutive hits, with a maximum of $414$ and a minimum of $165$. Under the varying-height setting, it achieved an average of $403.6$ hits, with a maximum of $462$ and a minimum of $283$. The numerical differences between the two settings are attributed to experimental noise, and both clearly demonstrate JuggleRL’s stable control performance and robust policy.

\begin{table}[h]
\centering
\caption{Real-world juggling performance under fixed and varying release heights}
\label{tab:real_world_main}
\begin{tabular}{ccc}
\toprule
\textbf{Height} & \textbf{Hits of 10 consecutive trials} & \textbf{Avg} \\
\midrule
\textbf{Fixed} & 301, 290, 325, 410, 401, 165, 217, \textbf{414}, 232, 355 & 311 \\
\textbf{Varying} & 454, 400, 460, 432, 460, 307, 355, 423, \textbf{462}, 283 & 403.6 \\
\bottomrule
\end{tabular}
\end{table}

Tab.~\ref{tab:performance_summary} compares JuggleRL with MBPP, where the MBPP results are taken from the original paper. It should be noted that the experimental setups differ: MBPP used higher toss heights to allow sufficient reaction time, whereas JuggleRL was evaluated without such constraints. JuggleRL significantly outperforms traditional methods, which achieved at most $14$ hits as reported in both \cite{eth} and~\cite{sjtu}. Videos of the juggling experiments are provided in the supplementary material.

\begin{table}[h]
\centering
\caption{Comparison of real-world hits between JuggleRL and MBPP}
\label{tab:performance_summary}
\begin{tabular}{lcc} 
\toprule
\textbf{Method} & \textbf{Average} & \textbf{Maximum} \\ 
\midrule
MBPP~\cite{eth} & 3.1 & 14.0 \\
JuggleRL-Fixed & 311.0 & 414.0 \\ 
JuggleRL-Varying & \textbf{403.6} & \textbf{462.0} \\ 
\bottomrule
\end{tabular}
\end{table}

\begin{figure}[h]
\centering
\includegraphics[width=0.95\columnwidth]{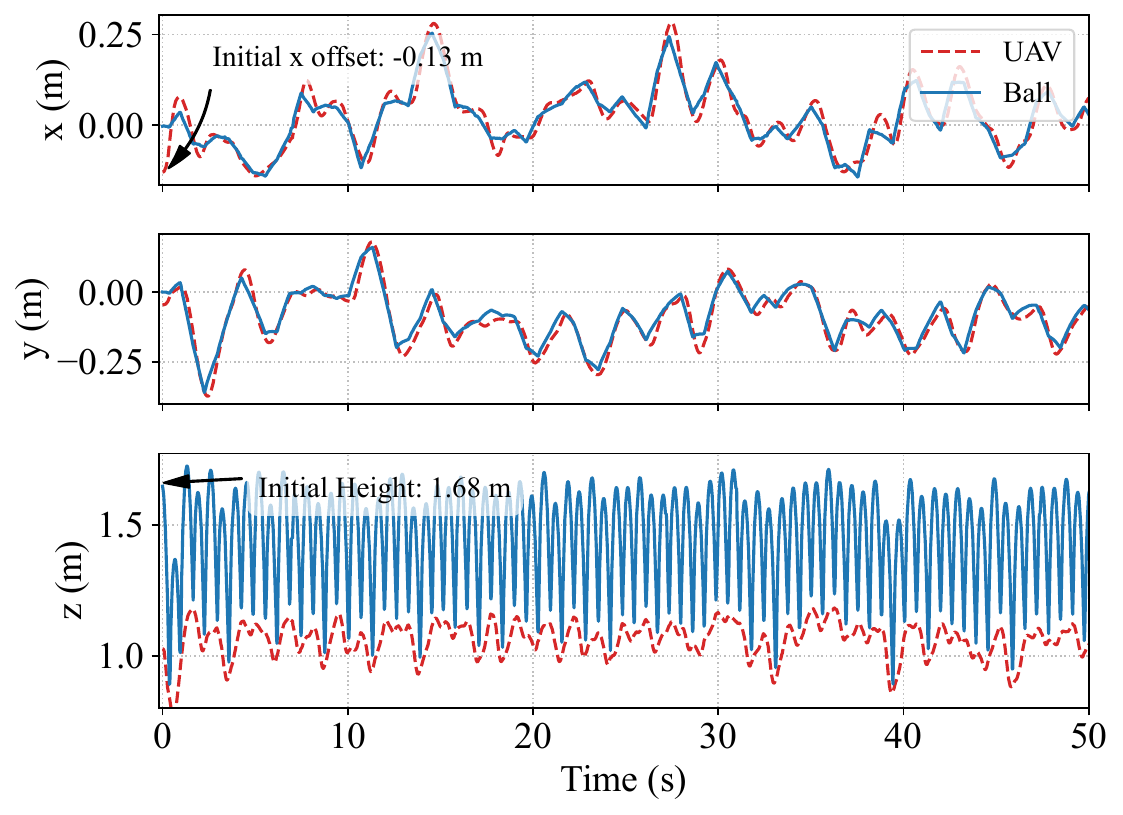}
\caption{Real-world juggling trajectory of a trial with a $0.13$ m horizontal offset, ball releasing from $1.68$ m.}
\label{fig:behavior}
\end{figure}

\begin{figure*}[h] 
\centering
\begin{minipage}{0.825\textwidth}
    \centering
    \includegraphics[width=\textwidth]{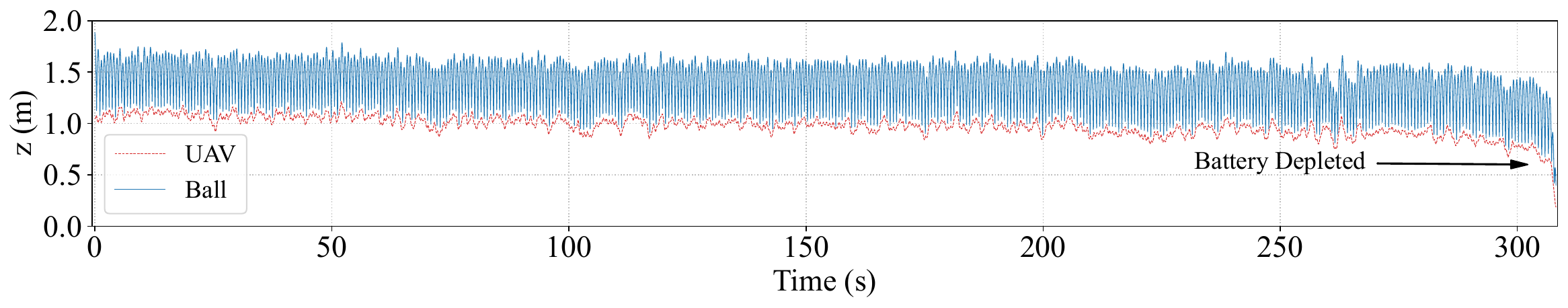}
    \subcaption{Z-axis trajectory}
    \label{fig:long_trajectory}
\end{minipage}
\begin{minipage}{0.16\textwidth}
    \centering
    \includegraphics[width=\textwidth]{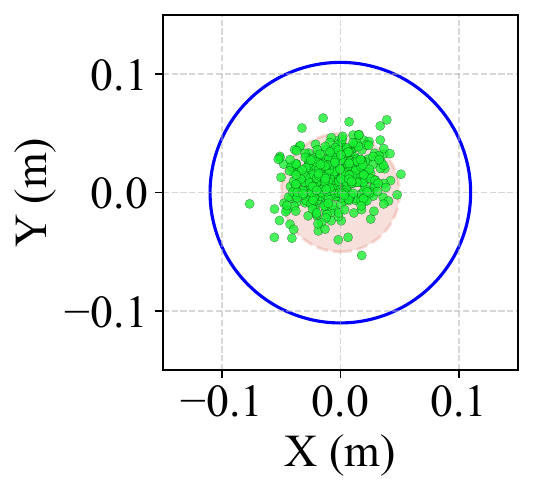} 
    \subcaption{Contact points}
    \label{fig:hitmap_add_to_wide}
\end{minipage}
\caption{Statistics of the longest trial (462 hits). (a) The gradual decrease in quadrotor height due to battery depletion. (b) Spatial distribution of contact points on the racket, with the ``sweet spot'' highlighted.}
\label{fig:combined_wide_figure}
\end{figure*}

\begin{figure*}[h]
    \centering
    \begin{subfigure}[b]{0.32\textwidth}
        \centering
        \includegraphics[height=4.5cm]{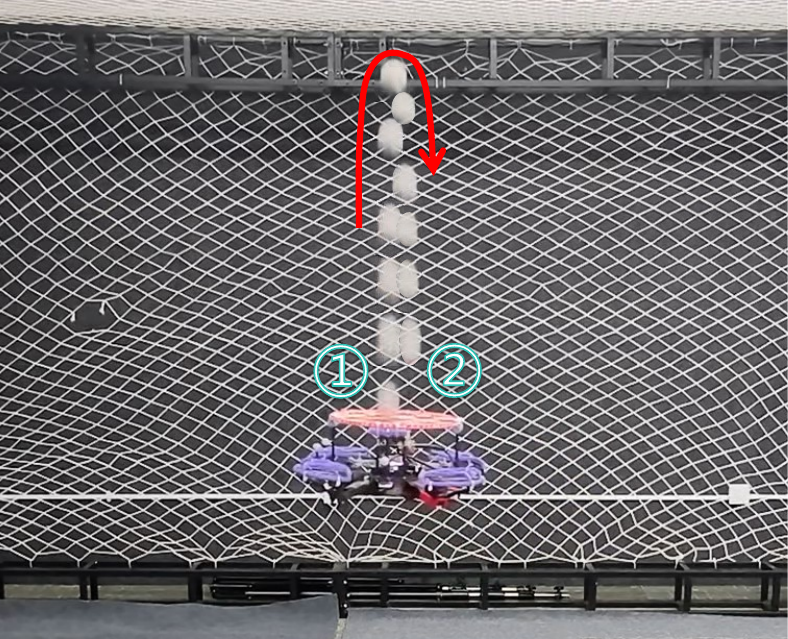} 
        \caption{Juggling near the origin}
        \label{fig:behavior_a}
    \end{subfigure}
    \begin{subfigure}[b]{0.32\textwidth}
        \centering
        \includegraphics[height=4.5cm]{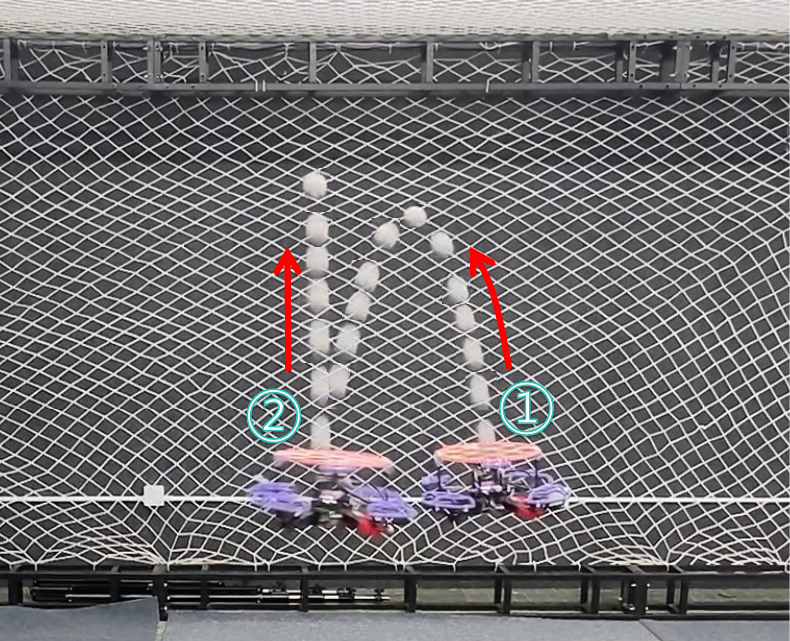}
        \caption{Directly correcting small deviations}
        \label{fig:behavior_b}
    \end{subfigure}
    \begin{subfigure}[b]{0.33\textwidth}
        \centering
        \includegraphics[height=4.5cm]{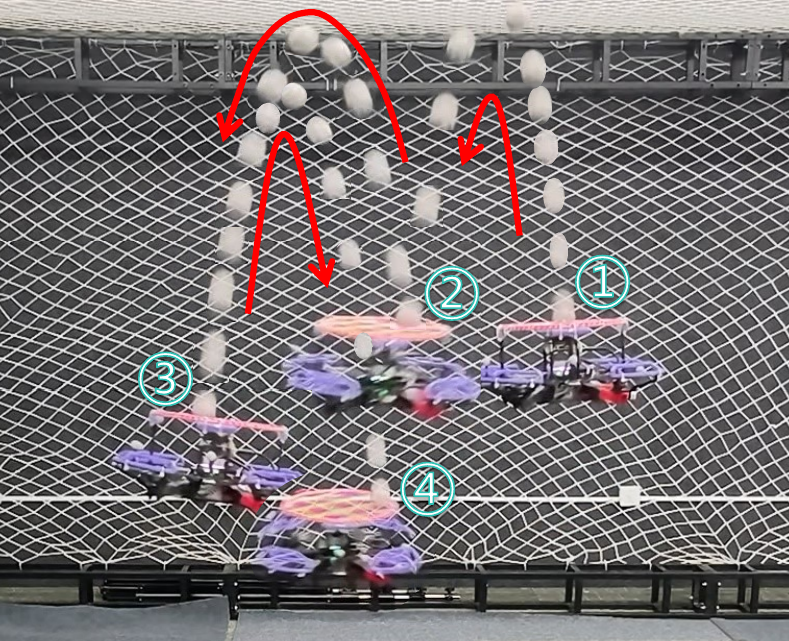}
        \caption{Two-stage ``stabilize-then-correct'' strategy}
        \label{fig:behavior_c}
    \end{subfigure}
    \caption{
        Time-lapse visualization of JuggleRL's adaptive strategies. (a) maintaining stable juggling near the origin; (b) directly correcting small deviations; and (c) adopting a two-stage ``stabilize-then-correct'' strategy when facing larger disturbances.
    }
    \label{fig:timelapse}
\end{figure*}

\textbf{Behavior Analysis.} In this part, we provide a detailed analysis of a single juggling trial.
\cref{fig:behavior} shows the $xyz$ trajectories of the quadrotor and the ball in a trial where the ball was released from a height of $1.68$ m. During hundreds of consecutive hits, the ball’s apex height remains consistently within the target range of $1.60$–$1.85$ m. Whenever a strike drives the ball outside this range, the quadrotor exhibits immediate dynamic adjustments that gradually restore stability. Moreover, the average horizontal displacement of the ball from the origin is only $0.129$ m. Considering the unavoidable deflections caused by the racket’s uneven surface~\cite{eth}, this highlights the strong closed-loop feedback regulation capability of JuggleRL.

\cref{fig:combined_wide_figure} presents the $z$-axis trajectory from the longest trial of $462$ consecutive hits. In the latter part of this trial, the quadrotor gradually descends due to battery depletion, and the experiment ultimately terminates when the battery is exhausted. The spatial distribution of contact points reveals that the policy predominantly strikes within the racket’s “sweet spot,” which reduces rebound unpredictability and further demonstrates the stability and reliability of JuggleRL.


\cref{fig:timelapse} illustrates the typical adaptive behaviors learned by the policy under different disturbance conditions: (a) maintaining stable juggling near the origin; (b) directly correcting small deviations; and (c) adopting a two-stage “stabilize-then-correct” strategy when facing larger disturbances. These behaviors suggest that JuggleRL does not rely on single-step optimal control. Instead, it achieves stability through gradual adjustments—a flexible control style particularly well suited for dynamic interaction tasks such as aerial juggling.

\subsection{Ablation Study}
We conduct ablation studies on the key components of JuggleRL through real-world experiments with a fixed release height. Tab. \ref{tab:ablation_study} summarizes the results.
 
\textbf{Reward shaping.}
To better understand the contribution of each reward component, we conduct ablation studies by removing them individually. Without the horizontal penalty (w/o $r^{\text{xy}}$), the quadrotor gradually drifts away from the origin and cannot maintain a centered position. Excluding the yaw penalty (w/o $r^{\text{yaw}}$) causes the quadrotor to spin aggressively, breaking flight stability. When the relative position reward (w/o $r^{\text{rpos}}$) is omitted, the quadrotor fails to follow the ball smoothly, instead showing jittery and erratic movements. Removing the contact reward (w/o $r^{\text{contact}}$) reduces the precision of strikes, leading to frequent off-center hits and more unpredictable rebounds. Finally, without the smoothness reward (w/o $r^{\text{smooth}}$), the quadrotor issues unstable commands, resulting in visibly shaky flight. These results collectively demonstrate that each component plays a critical role in achieving stable and robust juggling behavior.

\textbf{Domain randomization. }
Ablation studies on domain randomization highlight its critical role in robust juggling performance. When horizontal position randomization is removed (w/o H-Pos DR), the quadrotor struggles to compensate for misalignments, achieving only $0.5$ hits on average. Omitting restitution randomization (w/o Rest. DR) causes the agent to strike too aggressively, lowering the average hit count to $6.8$. Without height randomization (w/o Hgt. DR), the system cannot effectively adapt to variations in ball release elevation, resulting in unstable juggling and an average of $20.3$ hits. These results underscore that each randomization component is essential for enabling reliable real-world performance.

\textbf{LCP.} 
Removing the lightweight communication protocol (w/o LCP) reduces the average number of consecutive hits from about $311$ to $200.2$, while the standard deviation increases noticeably, indicating that communication delays amplify policy decision fluctuations and significantly reduce juggling stability. LCP mitigates high-frequency state estimation latency, improving observation quality and thereby ensuring more stable real-time control.



\begin{table}[h]
\centering
\renewcommand{\arraystretch}{1.5}  
\caption{Ablation study of key design components.}
\begin{tabular}{lcc}
\toprule
\textbf{Variants} & \textbf{Mean} & \textbf{Std Dev} \\
\midrule
JuggleRL & $\textbf{311.0}$ & $86.7$ \\
\hline \hline
w/o $r^{\text{xy}}$ & fail & / \\
w/o $r^{\text{yaw}}$ & fail & / \\
w/o $r^{\text{rpos}}$ & $58.7$ & $52.2$ \\
w/o $r^{\text{contact}}$ & $55.3$ & $56.0$ \\
w/o $r^{\text{smooth}}$ & $21.3$ & $19.6$ \\
\hline \hline
w/o H-Pos DR & $0.5$ & $0.6$ \\
w/o Rest. DR & $6.8$ & $5.2$ \\
w/o Hgt. DR & $20.3$ & $18.6$ \\
\hline \hline
w/o LCP & $200.2$ & $186.8$ \\
\bottomrule
\end{tabular}
\label{tab:ablation_study}
\end{table}

\begin{table}[htbp]
\centering
\caption{Real-world performance of JuggleRL on a lighter $5$g ball. The policy successfully adapts to the new ball weight with an average of $145.9$ hits across $10$ consecutive trials.}
\label{tab:lightweight_ball_results}
\setlength{\tabcolsep}{3pt} 
\begin{tabular}{cccccccccccc}
\toprule
\textbf{Trial} & $1$ & $2$ & $3$ & $4$ & $5$ & $6$ & $7$ & $8$ & $9$ & $10$ & Average \\ 
\midrule
\textbf{Hits} & $162$ & $23$ & $252$ & $494$ & $79$ & $92$ & $219$ & $77$ & $19$ & $42$ & $145.9$ \\ 
\bottomrule
\end{tabular}
\end{table}
\subsection{Generalization}
We further evaluate zero-shot transfer to an out-of-distribution setting: juggling a lighter $5$g ball. As shown in \cref{tab:lightweight_ball_results}, the policy achieves an average of $145.9$ hits over $10$ consecutive trials, with results ranging from $19$ to $494$. The high variance arises from the amplified sensitivity of the lighter ball to small control errors, yet the upper bound demonstrates that the learned strategy generalizes beyond the training distribution.



\section{Conclusion and Future Work}
We propose JuggleRL, a quadrotor-based ball juggling system designed for dynamic physical interaction. To enable robust sim-to-real transfer, the system integrates (i) high-fidelity simulation through system identification, (ii) carefully designed reward shaping and domain randomization, and (iii) a lightweight communication protocol that mitigates observation latency. With this design, JuggleRL achieves up to $462$ consecutive hits in real-world experiments, dramatically outperforming a model-based baseline that manages at most $14$ hits. In addition, the system demonstrates out-of-distribution generalization, successfully juggling a lighter $5$ g ball with an average of $145.9$ hits, despite its higher sensitivity to control errors.

Despite these advances, our system currently relies on an external motion capture setup, which limits autonomy. As future work, we aim to explore fully onboard, vision-based perception to further reduce external dependencies. In addition, we plan to extend the framework to multi-agent scenarios, such as cooperative juggling or aerial volleyball, where multiple drones coordinate in real time. These directions will further demonstrate the potential of reinforcement learning to enable agile, robust, and interactive aerial robotics.









\bibliographystyle{IEEEtran}
\bibliography{reference}

\end{document}